%% file: acl_latex.tex
\documentclass[11pt]{article}
\usepackage{bigfoot}
\usepackage[preprint]{acl}
\usepackage{amsmath}
\usepackage{cleveref}
\usepackage{times}
\usepackage{latexsym}
\usepackage{booktabs}
\usepackage{amsfonts} 
\usepackage{graphicx}
\usepackage{fontawesome5} % For beautiful icons
\usepackage[strings]{underscore}
\usepackage[T1]{fontenc}
\usepackage[utf8]{inputenc}

\usepackage{microtype}

\usepackage{inconsolata}

\usepackage{graphicx}
\newcommand\blfootnote[1]{%
  \begingroup
  \renewcommand\thefootnote{}\footnote{#1}%
  \addtocounter{footnote}{-1}%
  \endgroup
}

\title{Towards Intrinsic Interpretability of Large Language Models: \\ A Survey of Design Principles and Architectures}

\author{
  Yutong Gao\textsuperscript{1,3,*}, \quad
  Qinglin Meng\textsuperscript{4,*}, \quad 
  Yuan Zhou\textsuperscript{4}, \quad 
  Liangming Pan\textsuperscript{1,2,$\dagger$} \\
  \textsuperscript{1}MOE Key Lab of Computational Linguistics, Peking University \\
  \textsuperscript{2}Beijing Academy of Artificial Intelligence, Beijing, China \\
  \textsuperscript{3}Nanjing University of Science and Technology, \quad
  \textsuperscript{4}Purdue University \\
  \texttt{yutongkkk@njust.edu.cn}, \quad
  \texttt{\{meng160, zhou1475\}@purdue.edu} \\
  \texttt{liangmingpan@pku.edu.cn}
}

\begin{document}
\maketitle

\blfootnote{* Equal contribution.}
\blfootnote{$^\dagger$ Corresponding author.}

\begin{abstract}

While Large Language Models (LLMs) have achieved strong performance across many NLP tasks, their opaque internal mechanisms hinder trustworthiness and safe deployment. Existing surveys in explainable AI largely focus on post-hoc explanation methods that interpret trained models through external approximations. In contrast, intrinsic interpretability, which builds transparency directly into model architectures and computations, has recently emerged as a promising alternative. This paper presents a systematic review of the recent advances in intrinsic interpretability for LLMs, categorizing existing approaches into five design paradigms: functional transparency, concept alignment, representational decomposability, explicit modularization, and latent sparsity induction. We further discuss open challenges and outline future research directions in this emerging field. The paper list is available at: \href{https://github.com/PKU-PILLAR-Group/Survey-Intrinsic-Interpretability-of-LLMs}{\faGithub \ Survey-Intrinsic-Interpretability-of-LLMs}

\end{abstract}

\input{Introduction.tex}

\input{Two_Paradigms_in_Model_Interpretability}
\input{Design_Principles_of_Intrinsic_Interpretability}

\input{Intrinsic_Explanability_Methods}

\input{future_direction}

\bibliography{custom}

% \clearpage

\appendix
\section{Summary of Intrinsic Interpretability Methods}
\label{app:method_summary}
This appendix provides a consolidated summary of the intrinsic interpretability methods discussed throughout the survey. \Cref{tab:full_summary_comprehensive} organizes these approaches by design principle, highlighting their key mechanisms, sources of interpretability, and practical trade-offs in training cost, inference cost, and performance relative to black-box baselines. The table is intended as a reference for cross-method comparison rather than a comprehensive evaluation.

\begin{table*}[t]
\centering
\renewcommand{\arraystretch}{1.15}
\resizebox{\textwidth}{!}{
\begin{tabular}{@{}l l l l l l l@{}}
\toprule
\textbf{Method} & \textbf{Reference} & \textbf{Key Mechanism} & \textbf{Interp. Source} & \textbf{Train Cost} & \textbf{Infer. Cost} & \textbf{Perf.} \\
\midrule
\multicolumn{7}{c}{\textbf{Functional Transparency (Sec 4.1)}} \\
\midrule
GAMs & \citep{10.1214/ss/1177013604} & Additive smooth functions & Shape functions & Low & Low & Linear \\
GA$^2$M & \citep{Lou2013AccurateIM} & Pairwise interaction terms & Interaction maps & Medium & Low & Moderate \\
EBMs & \citep{nori2019interpretmlunifiedframeworkmachine} & Boosting for additive terms & Shape/Interaction & Medium & Low & Moderate \\
GAMI-Net & \citep{DBLP:journals/pr/YangZS21} & Neural shape functions & Individual NNs & Medium & Medium & Moderate \\
NODE-GAM & \citep{chang2022nodegamneuralgeneralizedadditive} & Neural subnetwork per feature & Shape functions & Medium & Medium & Moderate \\
SENN & \citep{alvarez2018towards} & Basis concepts + Relevance scores & Concepts \& Relevance & Medium & Medium & $\approx$ \\
B-cos Networks & \citep{bohle2022b} & Weight--input alignment transform & Linear explanations & Medium & Low & $\approx$ \\
B-cos LMs & \citep{wang2025b} & B-cos transforms + fine-tuning & Linear explanations & Medium & Low & $\approx$ \\
KANs & \citep{DBLP:conf/iclr/LiuWVRHS0T25} & Learnable splines on edges & 1D edge functions & High & High & TBD \\
Bilinear MLPs & \citep{DBLP:conf/iclr/PearceDROS25} & Bilinear interactions & Weight tensors & High & High & $\approx$ \\
\midrule
\multicolumn{7}{c}{\textbf{Concept Alignment (Sec 4.2)}} \\
\midrule
Standard CBMs & \citep{koh2020concept} & Hard concept bottleneck & Concept scores & Low & Low & $\downarrow$ \\
SCBMs & \citep{DBLP:conf/nips/VandenhirtzLMV24} & Joint concept distribution & Concept dependencies & Medium & Low & $\downarrow$ \\
Hybrid CBMs & \citep{DBLP:conf/nips/HavasiPD22, mahinpei2021promisespitfallsblackboxconcept} & Residual side-channel & Concepts + Residual & Low & Low & $\approx$ \\
CB-LLM & \citep{DBLP:conf/iclr/SunOUW25} & Hybrid bottleneck + Adversarial & Concepts + Latent & High & Low & $\approx$ \\
Label-free CBM & \citep{DBLP:conf/iclr/OikarinenDNW23} & Auto-discovery via CLIP & Concept scores & Medium & Low & $\downarrow$ \\
CEMs / IntCEMs & \citep{DBLP:conf/nips/ZarlengaBCMGDSP22} & Concept embeddings & Concept Vectors & Medium & Low & $\approx$ \\
Codebook Features & \citep{DBLP:conf/icml/TamkinTG24} & Vector Quantization (VQ) & Discrete Codes & Medium & Low & $\downarrow$ \\
\midrule
\multicolumn{7}{c}{\textbf{Representational Decomposability (Sec 4.3)}} \\
\midrule
Backpack & \citep{DBLP:conf/acl/HewittTML23} & Sense vectors + Context weights & Sense vectors & Medium & High & $\downarrow$ \\
Char-BLM & \citep{sun-hewitt-2023-character} & Character-level sense vectors & Character senses & Medium & High & $\downarrow$ \\
CoCoMix & \citep{tack2025llmpretrainingcontinuousconcepts} & Concept prediction \& mixing & Continuous concepts & Medium & High & $\approx$ \\
\midrule
\multicolumn{7}{c}{\textbf{Explicit Modularization (MoEs) (Sec 4.4)}} \\
\midrule
\multicolumn{7}{l}{\textit{--- Intra-Expert Sparsity ---}} \\
MoE-X & \citep{DBLP:conf/icml/YangVKWDB025} & Sparsity-aware routing + ReLU & Sparse Experts & Low & Low & $\approx$ \\
MoV & \citep{DBLP:conf/iclr/ZadouriUAELH24} & Mixture of Vectors (Linear) & Linear Units & Low & Low & $\approx$ \\
MoLORA & \citep{DBLP:conf/iclr/ZadouriUAELH24} & Mixture of LoRA adapters & Low-rank Adapters & Low & Low & $\approx$ \\
\multicolumn{7}{l}{\textit{--- Fine-Grained Decomposition ---}} \\
MONET & \citep{DBLP:conf/iclr/ParkJKK25} & Product Key Composition & Monosemantic Exp. & High & Medium & $\approx$ \\
MxD & \citep{oldfield2025interpretabilitysacrificefaithfuldense} & Tensor factorization & Linear sublayers & High & Medium & $\approx$ \\
MPO-MoE & \citep{gao-etal-2022-parameter} & Matrix Product Operator & Shared tensors & Medium & Medium & $\approx$ \\
\multicolumn{7}{l}{\textit{--- Semantically Aligned Routing ---}} \\
Task-Based MoE & \citep{pham2023taskbasedmoemultitaskmultilingual} & Task embeddings + Adapters & Task Adapters & Low & Low & $\uparrow$ \\
Lingual-SMoE & \citep{zhao2024sparse} & Language-guided routing & Language Experts & Low & Low & $\uparrow$ \\
THOR-MoE & \citep{liang2025thormoehierarchicaltaskguidedcontextresponsive} & Hierarchical Context-Routing & Domain Experts & Medium & Low & $\uparrow$ \\
Apollo-MoE & \citep{DBLP:conf/iclr/ZhengWL0ZW25} & Language Family Grouping & Family Experts & Medium & Low & $\uparrow$ \\
RoMA & \citep{li2025routingmanifoldalignmentimproves} & Manifold Alignment Reg. & Routing Geometry & Medium & None & $\uparrow$ \\
USMoE & \citep{do2025unifiedsparsemixtureexperts} & Linear Programming Routing & Unified Scores & Low & Low & $\uparrow$ \\
Orthogonality & \citep{guo2025advancingexpertspecializationbetter} & Orthogonality \& Variance loss & Exclusive Experts & Low & Low & $\approx$ \\
\midrule
\multicolumn{7}{c}{\textbf{Latent Sparsity Induction (Sec 4.5)}} \\
\midrule
Weight-Sparse & \citep{gao2025weightsparsetransformersinterpretablecircuits} & $L_0$ Regularization & Sparse Circuits & Very High & Low & $\downarrow$ \\
GLUs / SwiGLU & \citep{DBLP:conf/icml/DauphinFAG17, shazeer2020gluvariantsimprovetransformer} & Gated Linear Units & Activation Paths & Low & Low & $\uparrow$ \\
\bottomrule
\end{tabular}
}
\caption{A comprehensive summary of intrinsic interpretability architectures covering all methods discussed in this survey. \textbf{Perf.} denotes reported performance vs. black-box baselines: $\uparrow$ (Improved), $\approx$ (Similar), $\downarrow$ (Trade-off). Note: KANs and Bilinear MLPs are categorized under Functional Transparency following the text structure.}
\label{tab:full_summary_comprehensive}
\end{table*}

\end{document}

%% file: Introduction.tex
\section{Introduction}

Large Language Models have achieved remarkable success across diverse tasks \citep{brown2020languagemodelsfewshotlearners,DBLP:journals/jmlr/RaffelSRLNMZLL20,chowdhery2022palmscalinglanguagemodeling,geminiteam2025geminifamilyhighlycapable}. However, their complexity often makes them "black boxes" \citep{bommasani2022opportunitiesrisksfoundationmodels}, hiding their internal decision-making. This lack of transparency creates trust and safety risks, especially in high-stakes fields like healthcare and law \citep{DBLP:journals/natmi/Rudin19,DBLP:conf/cybersa/PawarORO20}.

To address these concerns, interpretability research is often divided into two paradigms: post-hoc explanation and intrinsic design. Post-hoc methods analyze trained, fixed models using external tools such as LIME, SHAP, sparse autoencoders, or causal interventions \citep{DBLP:conf/kdd/Ribeiro0G16,DBLP:conf/nips/LundbergL17,DBLP:conf/iclr/HubenCRES24,DBLP:conf/nips/MengBAB22}. Many rely on surrogate models or statistical attributions, resulting in a well-known fidelity gap between the explanation and the model’s true computation \citep{DBLP:conf/acl/JacoviG20}.
Causal based post hoc methods partially address this issue by intervening directly on internal components, yielding stronger local faithfulness \citep{DBLP:conf/nips/MengBAB22,DBLP:conf/iclr/WangVCSS23}. However, their explanations remain highly fine grained and are difficult to aggregate into coherent, high level accounts of overall model behavior.

In contrast, intrinsic interpretability builds transparency directly into the model architecture and training process \citep{DBLP:journals/jmlr/FedusZS22,gao2025weightsparsetransformersinterpretablecircuits}. By ensuring that the model’s internal computation is itself interpretable, these approaches aim to achieve \textit{structural fidelity}, namely a direct correspondence between model behavior and its explanation, without relying on external surrogates or post-hoc aggregation. Historically, however, intrinsic methods were constrained by a severe trade-off: models that were transparent by construction typically lacked the expressive power required for complex language tasks \citep{DBLP:journals/entropy/LinardatosPK21}.

Recent advances demonstrate that interpretability and performance need not be mutually exclusive, showing that large-scale models can be designed with interpretable internal structure while retaining competitive task performance
\citep{DBLP:journals/natmi/Rudin19,DBLP:journals/tmlr/SharkeyCBLWBGHOBBGCNRWS25}. By incorporating inductive biases such as modularity, sparsity, disentanglement, and structured representations directly into modern architectures and training objectives \citep{DBLP:conf/iclr/ShazeerMMDLHD17,DBLP:conf/iclr/LouizosWK18,DBLP:journals/jmlr/FedusZS22,gao2025weightsparsetransformersinterpretablecircuits}, these methods enable interpretability to emerge as a property of the model itself rather than as an after-the-fact analysis.

Despite this rapid progress, the literature on intrinsic interpretability remains fragmented, spanning disparate model classes, architectural choices, and training principles. Unlike post-hoc explanation methods whose taxonomy and limitations have been extensively surveyed \citep{molnar2025,Madsen_2022,DBLP:journals/tist/ZhaoCYLDCWYD24,palikhe2025transparentaisurveyexplainable}, there remains a need for a unified framework that organizes intrinsic approaches around shared design principles or clarifies how different mechanisms contribute to transparency in LLMs. This survey aims to fill this gap by systematically reviewing intrinsic interpretability methods for LLMs, distilling common design principles, and highlighting open challenges and promising future directions. 

Our contributions are threefold. First, we distinguish post-hoc explanation from intrinsic interpretability, clarifying their differences in faithfulness, scope, and design philosophy. Second, we introduce a structured taxonomy of intrinsic interpretability methods organized around five core design principles: \textit{Functional Transparency}, \textit{Concept Alignment}, \textit{Representational Decomposability}, \textit{Explicit Modularity}, and \textit{Latent Sparsity Induction}. Finally, we synthesize existing work within this framework, analyze methodological strengths and limitations, and identify key open challenges and future research directions.

%% file: Two_Paradigms_in_Model_Interpretability.tex
\section{Two Paradigms of LLM Interpretability}
\begin{figure}[ht]
    \centering
    \includegraphics[width=0.95\linewidth]{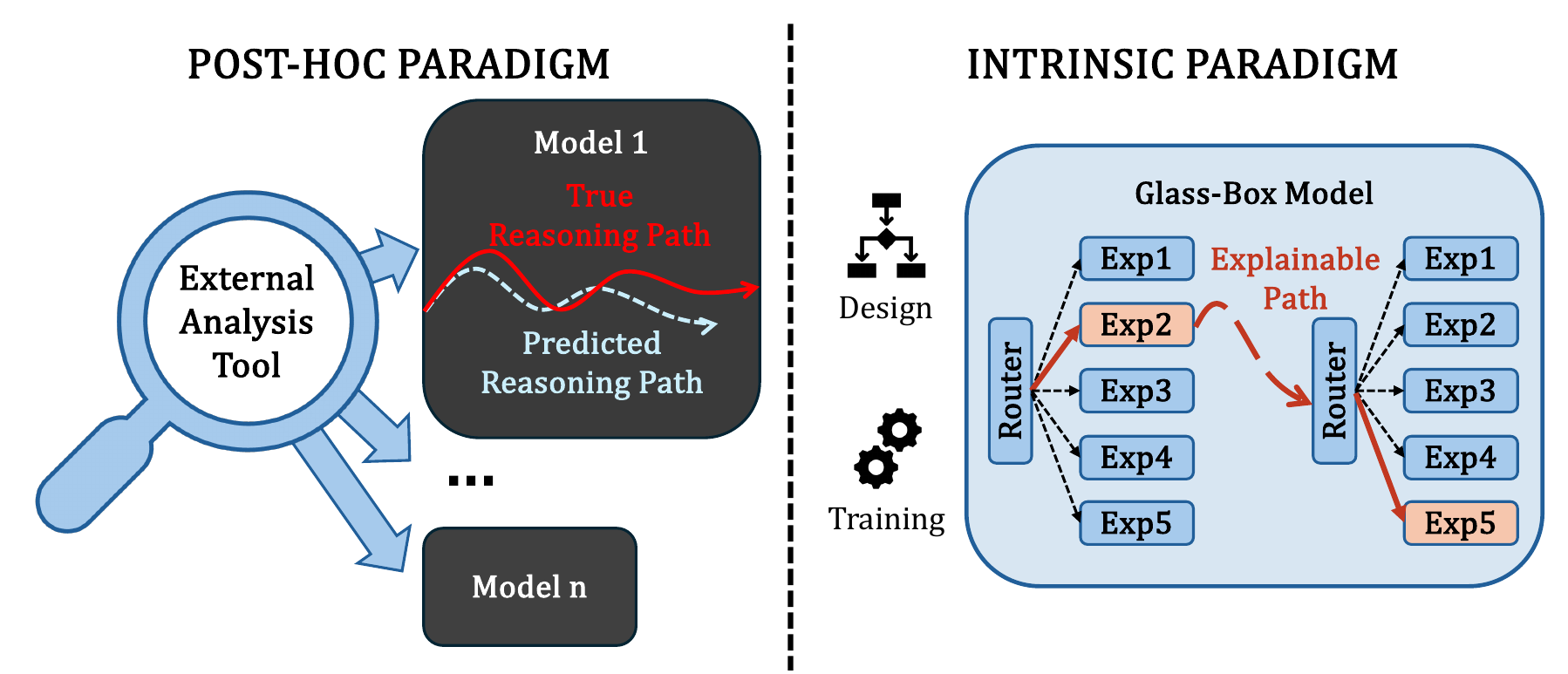}
    \caption{Comparison of the Post-hoc analysis versus Intrinsic design in LLM interpretability.}
\label{fig:paradigm_comparison}
    % \label{fig:placeholder}
\end{figure}

Research on interpretability for modern neural models has largely converged around two paradigms: 
(1) \textit{post-hoc analysis}, which applies external tools to a trained, fixed model, and 
(2) \textit{intrinsic interpretability}, which incorporates transparency directly into the model's architecture and training process. 
We distinguish these paradigms by causal necessity: an interpretability method is intrinsic if its interpretable components (e.g., sparse experts or concepts) lie on the critical computation path, such that modifying them directly alters the model's output. While recent hybrid approaches \citep{DBLP:conf/nips/HavasiPD22, tack2025llmpretrainingcontinuousconcepts} blur this line by allowing information to flow through residual side channels to preserve performance, we classify them as intrinsic designs that trade partial structural fidelity for enhanced capability. \Cref{fig:paradigm_comparison} summarizes the key conceptual differences between these two approaches.

\subsection{Post-hoc Interpretability}

Post-hoc analysis has long dominated interpretability research as the default approach for explaining complex neural models. Existing surveys extensively cover these methods, ranging from early feature attribution techniques to modern mechanistic and causal analyses \citep{Madsen_2022,DBLP:journals/tist/ZhaoCYLDCWYD24,ji2025aialignmentcomprehensivesurvey,palikhe2025transparentaisurveyexplainable}.

Most post-hoc methods operate at one of two levels. At the \textit{behavioral level}, feature attribution techniques such as LIME and SHAP \cite{DBLP:conf/kdd/Ribeiro0G16,DBLP:conf/nips/LundbergL17} estimate input importance by perturbing inputs and observing output changes, treating the model largely as a black box. At the \textit{internal level}, inspection methods analyze intermediate representations. Probing classifiers train external predictors to detect concepts in hidden states \citep{DBLP:journals/jmlr/RaffelSRLNMZLL20}, while LogitLens projects hidden representations into the vocabulary space to expose transient computations \citep{nostalgebraist2020logitlens}. More recently, SAEs have emerged as a mechanistic tool for decomposing polysemantic activations into sparse, interpretable features \citep{DBLP:conf/iclr/HubenCRES24}.

Despite their flexibility, post-hoc methods share a fundamental limitation: they rely on auxiliary approximations rather than the model’s native computation. \cite{DBLP:conf/acl/JacoviG20} Attribution methods depend on local surrogate models, probing approaches identify correlations without establishing causal use \citep{ravichander-etal-2021-probing}, and mechanistic tools such as SAEs introduce reconstruction error by approximating, rather than exactly reproducing, forward-pass activations.

Causal-based post-hoc methods partially mitigate these issues by intervening on internal components and measuring their effects on model outputs \cite{DBLP:conf/nips/MengBAB22, DBLP:conf/iclr/WangVCSS23}. While such interventions provide stronger local faithfulness, their fine-grained nature makes it difficult to aggregate localized causal effects into coherent, high-level explanations of overall model behavior.

\subsection{Intrinsic Interpretability}
Intrinsic interpretability addresses the fidelity gap by designing models whose internal computation is transparent by construction. Rather than analyzing a trained black-box model, intrinsic approaches aim to build models in which the explanation is inseparable from the computation itself. As a result, interpretability is achieved without relying on post-hoc approximations.

Historically, intrinsic interpretability was largely confined to simple and low-dimensional models, such as linear regressors or generalized additive models, whose transparency comes at the cost of limited expressive power \citep{nelder1972generalized,10.1214/ss/1177013604,DBLP:journals/entropy/LinardatosPK21}. While effective for certain tasks, these models were insufficient for complex NLP tasks. %A central challenge—and the focus of this survey—is extending intrinsic interpretability to large-scale neural architectures. 
However, recent progress in sparse modeling, modular architectures, and structured representations suggests that transparency and scalability need not be mutually exclusive, enabling intrinsically interpretable designs that retain competitive performance at scale \citep{DBLP:journals/jmlr/FedusZS22,gao2025weightsparsetransformersinterpretablecircuits,DBLP:conf/icml/TamkinTG24}. The following sections present the core design principles in \Cref{sec:principles} and representative methods in \Cref{sec:methods} underlying this line of work.

\begin{figure*}[t]
    \centering
    \includegraphics[width=0.98\textwidth]{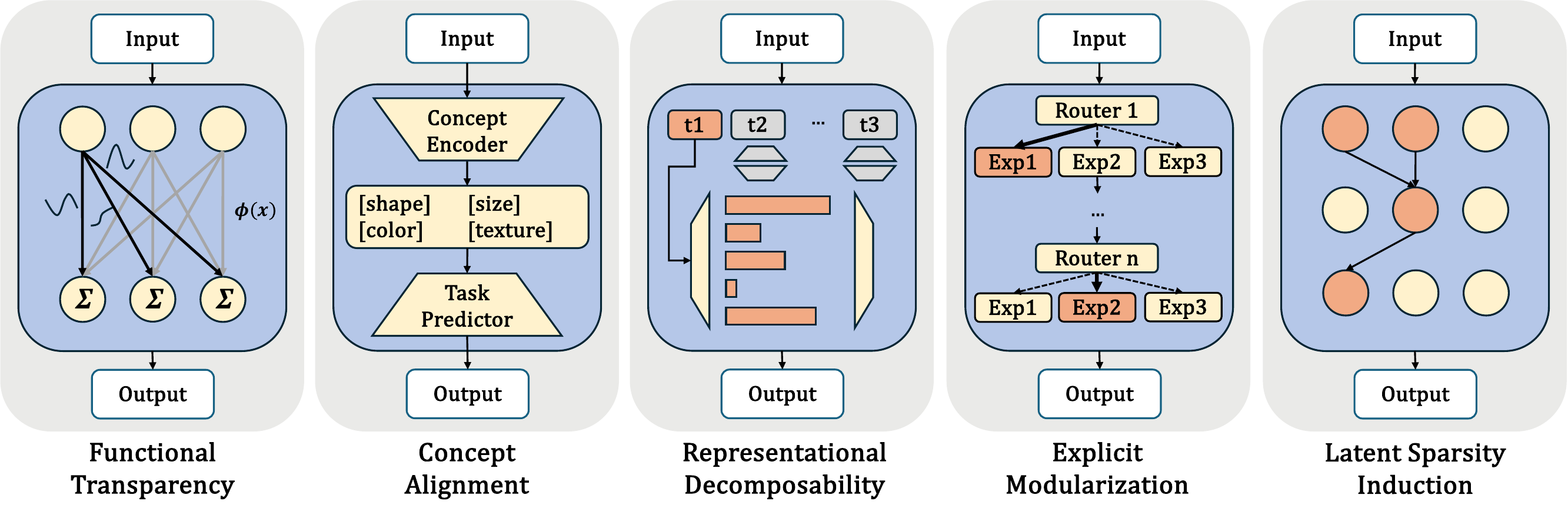}
    \caption{\textbf{A taxonomy of intrinsic architectural designs for interpretable LLMs.} We categorize existing approaches into five primary families based on their core mechanism for transparency.}
    \label{fig:foundational_ideas}
\end{figure*}

%% file: Design_Principles_of_Intrinsic_Interpretability.tex
\section{Design Principles of Intrinsic Interpretability}
\label{sec:principles}

As illustrated in Figure~\ref{fig:foundational_ideas}, we categorize intrinsic interpretability into five design principles. These design philosophies dictate \textit{how} transparency is constructed within a model. In this section, we analyze the rationale, formulation, and trade-offs of each principle, connecting them to the specific methodologies detailed in \Cref{sec:methods}.
\paragraph{Functional Transparency.}
This principle advocates architectures whose computations are both structurally explicit and semantically meaningful. Rather than relying on opaque compositions of dense layers, such models are organized so that both the \textit{where} (through structured or decomposed components) and the \textit{what} (through operations with clear mathematical semantics) of computation are directly inspectable. As a result, these models behave less like black boxes and more like readable algorithms. Representative implementations are discussed in \Cref{subsec:functional transparency}. Key trade-offs of this approach include reduced expressivity and training efficiency.

\paragraph{Concept Alignment.}
While functional transparency emphasizes mathematical structure, concept alignment targets semantic interpretability. This principle encourages latent variables to correspond directly to human-understandable concepts %(e.g., $z_k \approx \text{concept}_k$)
, thereby reducing \textit{polysemanticity}, where individual units encode multiple unrelated features. By aligning representations with explicit concepts, models become easier to interpret and reason about. The primary trade-off is an \textit{alignment tax}: constraining representations to be human-interpretable may limit expressive capacity or require additional supervision. Representative approaches following this principle are discussed in \Cref{subsec:alignment}.

\paragraph{Representational Decomposability.}
Extending alignment, this principle focuses on the geometry of the latent space. It seeks to disentangle representations into independent subspaces so that distinct factors of variation can be manipulated separately without interference. This separation enables more precise and controllable generation. The central challenge is enforcing such decomposability, for example through orthogonality constraints, without relying on extensive supervision or sacrificing flexibility. Recent architectures that instantiate this principle are reviewed in \Cref{subsec:representation decomp}.

\paragraph{Explicit Modularization.}
Whereas traditional models operate as a single monolithic block, this principle advocates decomposing computation into distinct, independently functioning modules. A routing mechanism explicitly selects which modules process a given input, yielding a clear and traceable computational pathway. A prominent instantiation of this principle is the Mixture-of-Experts (MoE) architecture (Section~\ref{subsec:moe}), which introduces transparency by structuring the model around specialized functional units. A key trade-off of this approach is the added complexity of routing and coordination, which can complicate optimization and limit global expressivity.

\paragraph{Latent Sparsity Induction.}
Rather than imposing a hand-crafted modular structure, this principle aims to induce modularity within otherwise standard neural architectures. The core insight is that the opacity of dense networks often stems from uniformly active and highly entangled pathways. Selective activation can be encouraged through sparsity-inducing training objectives, such as $L_0$ or structured regularization, or through competitive gating mechanisms such as Gated Linear Units (GLUs), which conditionally route information. These mechanisms encourage the model to suppress redundant channels and form task-specific subcircuits. Representative techniques following this principle are discussed in \Cref{subsec:latent sparsity induction}. A key trade-off is that strong sparsity or gating constraints can complicate optimization and may reduce expressivity or robustness if not carefully tuned.

We emphasize that these five paradigms serve as organizing design principles rather than strictly disjoint categories. Since some methods may instantiate multiple principles, we classify them according to the mechanism that most directly embeds interpretability into the model’s architecture, training objective, or primary computation path, while noting cross-category connections where appropriate.

%% file: Intrinsic_Explanability_Methods.tex
\section{Intrinsic Interpretability Methods}\label{sec:methods}
\begin{figure}
    \centering
    \includegraphics[width=0.95\linewidth]{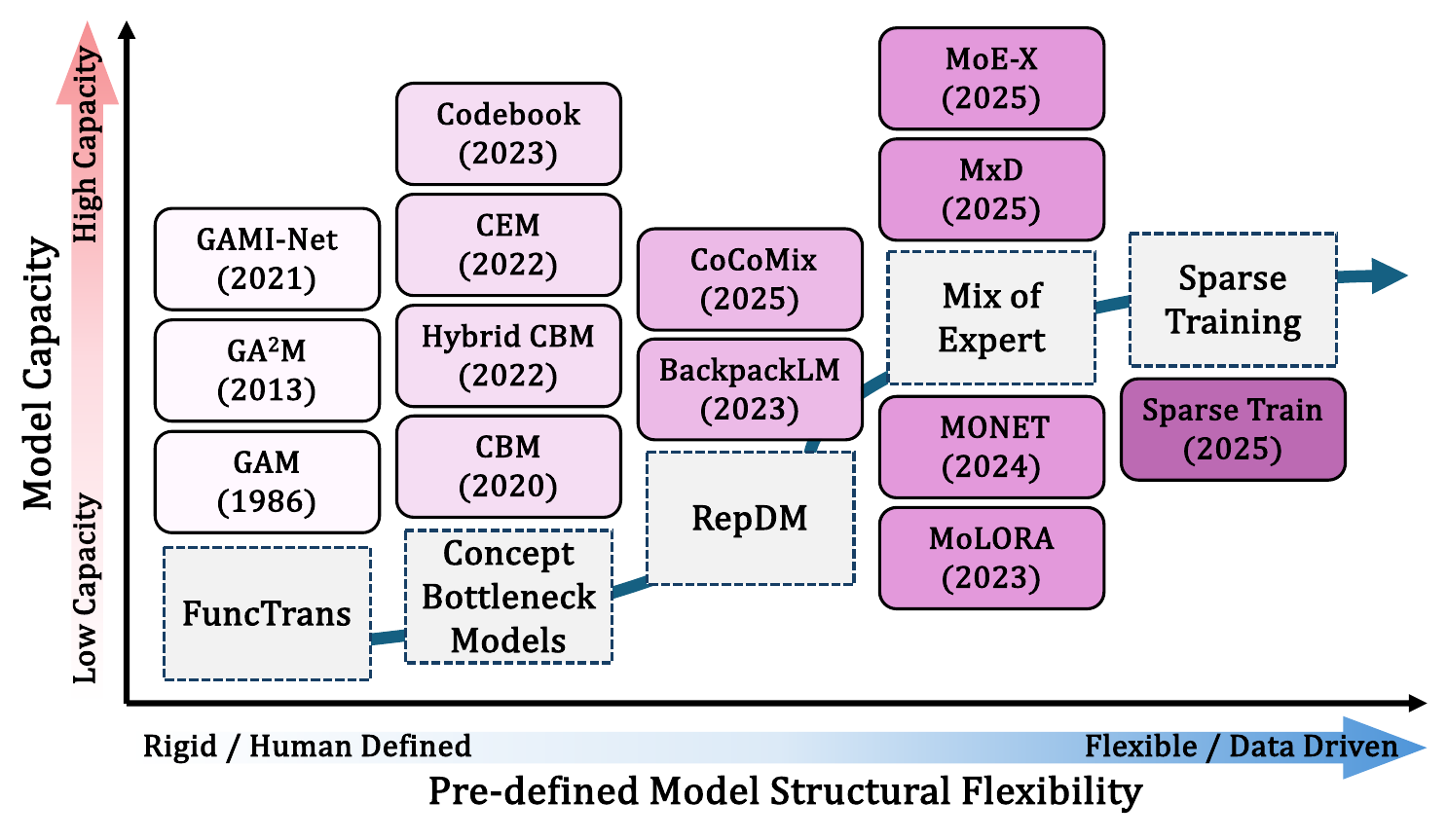}
    \caption{\textbf{Evolution of intrinsic interpretability.} The field has shifted from rigid, human-defined structures (e.g., GAMs) to scalable, data-driven sparse architectures (e.g., Specialized MoEs) that balance interpretability with performance.}
    \label{fig:timeline}
\end{figure}
In this section, we organize existing intrinsic interpretability methods according to the design principles introduced in \Cref{sec:principles}. \Cref{fig:timeline} provides an overview of the methods discussed in this section, situating them along two dimensions: structural flexibility and model capacity.

\subsection{Functional Transparency}
\label{subsec:functional transparency}

In this subsection, we introduce three representative model families that realize functional transparency through architectural design, progressing from simple to more complex structures. 

\paragraph{Generalized Additive Models.}
GAMs were originally proposed by \citet{10.1214/ss/1177013604} as an extension of GLMs \citep{nelder1972generalized}. Instead of modeling the response as a linear combination of input features, GAMs replace the linear predictor with a sum of smooth univariate functions, yielding the formulation
\[
F(\mathbf{x}) = f_0 + \sum_i f_i(x_i)
\]
where each $f_i$ is a learned smooth function of a single feature. These functions are typically estimated using iterative backfitting or local scoring procedures, which preserve interpretability by keeping each feature’s contribution explicit and separable.

To capture limited feature interactions while retaining interpretability, \citet{Lou2013AccurateIM} introduced $\mathrm{GA^2M}$, defined as\[F(\mathbf{x}) = f_0 + \sum_i f_i(x_i) + \sum_{i,j} f_{ij}(x_i, x_j)\]
While effective, modeling pairwise interactions significantly increases computational and statistical complexity. This challenge was later addressed by EBMs \citep{nori2019interpretmlunifiedframeworkmachine}, which use modern boosting techniques to efficiently learn additive and low-order interaction terms.

\paragraph{Neural Additive Models.}
More recently, researchers have leveraged neural networks to replace the smooth functions in GAMs, increasing expressivity while preserving the additive structure. Representative examples include GAMI-Net \citep{DBLP:journals/pr/YangZS21}, as well as NODE-GAM and NODE-$\mathrm{GA^2M}$ models \citep{chang2022nodegamneuralgeneralizedadditive}. In these approaches, each feature (or feature pair) is modeled by a small neural subnetwork, enabling nonlinear function approximation while maintaining per-feature transparency and interpretability. 
% \textcolor{red}{TODO: Lack of scalability}

\paragraph{Self-Explaining Neural Networks.}
Beyond additive models, \citet{alvarez2018towards} proposed Self-Explaining Neural Networks (SENN), which build predictions from an explicit combination of interpretable basis concepts and corresponding relevance scores. Concretely, SENN generalizes linear models by allowing both the concepts and their coefficients to be learned functions of the input, while imposing stability and interpretability constraints so that the contribution of each concept to the final prediction remains transparent.

\paragraph{B-cos Networks}
Another line of work modifies the predictive transformation itself to better align model parameters with task-relevant evidence. \citet{bohle2022b} proposed B-cos Networks, which replace standard linear transformations with B-cos transforms that encourage weight--input alignment during training. A B-cos transformer is defined as
\[
\text{B-cos}(x;w) = \|\hat{w}\|\|x\| |c(x,\hat{w})|^{B} \times \text{sgn}(c(x,\hat{w}))
\] where $c(x,\hat{w}) = \text{cos}(\angle(x,w))$ As a result, the overall computation yields linear explanations that more faithfully reflect the evidence used for prediction. More recently, this idea has been extended to language models through B-cos LMs \citep{wang2025b}, which adapt pretrained language models with B-cos style transformations and fine-tuning to improve explanation faithfulness in NLP settings.

\paragraph{Kolmogorov--Arnold Networks.}
A more radical departure from the standard perceptron architecture is the Kolmogorov--Arnold Network (KAN) \citep{DBLP:conf/iclr/LiuWVRHS0T25}. While traditional multilayer perceptrons place learnable weights on edges and fixed activation functions on nodes, KANs invert this design by assigning learnable univariate functions, often parameterized as splines, to the edges. Based on the Kolmogorov--Arnold representation theorem, a KAN represents a multivariate function as

\[f(\mathbf{x}) = \sum_{q=1}^{2n+1} \Phi_q \left( \sum_{p=1}^{n} \phi_{q,p}(x_p) \right)\]
This formulation offers a high degree of functional transparency, as each $\phi_{q,p}$ can be directly visualized as a one-dimensional curve. As a result, KANs are relatively \emph{symbolic-friendly}: in some cases, trained networks can be pruned and further simplified via symbolic regression into concise mathematical expressions. However, \citet{hou2025kolmogorovarnoldnetworkscriticalassessment} showed that KANs often suffer from significant computational overhead, optimization instability and inferior performance compared to standard MLPs when model size or input dimensionality grows.

\subsection{Concept Alignment} \label{subsec:alignment}
Concept alignment is primarily realized through CBMs, which enforce interpretability by structurally constraining information flow within the network. Unlike post-hoc probes that analyze fixed representations, intrinsic CBMs explicitly design the architecture as a composition of a concept encoder $g: \mathcal{X} \rightarrow \mathcal{C}$ and a predictor $f: \mathcal{C} \rightarrow \mathcal{Y}$

\paragraph{Standard CBMs (Hard Bottlenecks).}
First formalized by \citet{koh2020concept}, standard CBMs impose a strict bottleneck where the final prediction relies exclusively on the predicted concepts $\hat{c} = g(x)$. This can be achieved via \textit{independent training} (training $g$ and $f$ sequentially) or \textit{joint training}. 
\citet{DBLP:conf/nips/VandenhirtzLMV24} proposed SCBMs to relax the assumption that concepts are conditionally independent by learning a joint distribution over concept rather than predicting each concept separately.
While this architecture guarantees that the reasoning process is grounded in the defined concepts, it often suffers from an accuracy-interpretability trade-off, as the bottleneck may discard task-relevant information not captured by the predefined concept set.

\paragraph{Hybrid CBMs.}
To mitigate the performance degradation of hard bottlenecks, \citet{mahinpei2021promisespitfallsblackboxconcept} and \citet{DBLP:conf/nips/HavasiPD22} proposed Hybrid CBMs. These models introduce a side channel, allowing the predictor to access both the explicit concepts $c$ and uncontrolled latent embeddings $z$ (i.e., $y = f(c, z)$). 
CB-LLM \cite{DBLP:conf/iclr/SunOUW25} extends this hybrid paradigm to LLMs, which introduces an unsupervised latent pathway alongside the concept bottleneck and employs adversarial training to remove concept-related information from the latent channel.
Interpretability is maintained by applying regularization during training to maximize the model's reliance on concept while using $z$ without encoding concept-related information.

\paragraph{Concept Embedding Models (CEMs).}
In NLP tasks, compressing a concept to a single scalar activation limits expressivity. To address this issue, CEMs and IntCEMs \citep{DBLP:conf/nips/ZarlengaBCMGDSP22} represent each concept as a high-dimensional vector in a learnable subspace rather than a scalar. This design allows the model to capture nuances (e.g., polysemy) while strictly restricting the downstream predictor to linear interactions between these concept embeddings, preserving the distinct attribution of the bottleneck design. 

\paragraph{Unsupervised Discrete Bottlenecks.}
A limitation of the preceding approaches is their reliance on predefined concept annotations. To address this, \citet{DBLP:conf/icml/TamkinTG24} proposed Codebook Features, which introduce an intrinsic bottleneck in a fully unsupervised manner. The method applies vector quantization \citep{Gray1984VectorQ,DBLP:conf/nips/OordVK17} to approximate continuous hidden states using sparse combinations of vectors from a learned codebook, trained by jointly optimizing the language modeling objective and a reconstruction loss. By restricting representations to a discrete vocabulary, the approach promotes the emergence of distinct, often human-interpretable features without manual annotation. However, empirical results are reported on relatively small language models and a limited set of tasks, leaving its behavior at larger scales an open question.

\subsection{Representational Decomposability Models}
\label{subsec:representation decomp}

This class of methods operationalizes this design principle by explicitly structuring the model’s latent space. Unlike standard Transformer architectures, where information is distributed across a single dense hidden representation, these approaches impose geometric constraints that separate distinct semantic factors into orthogonal subspaces or parallel processing streams.

\paragraph{Backpack Language Models.}
Standard Transformers entangle contextual information and lexical identity within a unified hidden state, making it difficult to isolate the contribution of individual word senses. To address this limitation, \citet{DBLP:conf/acl/HewittTML23} propose the Backpack Language Model (BLMs), which decomposes prediction into interpretable components. In this architecture, each vocabulary item is associated with a set of learnable, non-contextual \textit{sense vectors}, capturing different meanings of the same surface form.
% (e.g., ``apple'' as a fruit versus ``Apple'' as a company). 
The self-attention mechanism is constrained to produce non-negative weights, which combine these sense vectors additively:
\[
    y = \text{Unembed}\left( \sum_{i=1}^{n} \alpha_i(\mathbf{x}) \cdot v_{\text{sense}}^{(i)} \right)
\]
By construction, the output representation is a weighted sum of independent sense vectors, enabling direct inspection and targeted intervention. Subsequent work extends this framework to non-alphabetic languages via Character-level Chinese BLMs \citep{sun-hewitt-2023-character}, which learn interpretable sense decompositions at the character level, as well as to downstream control tasks, including model editing via canonical examples, where modifying or fine-tuning specific sense vectors enables localized behavioral changes without broadly perturbing the model \citep{hewitt2024modeleditingcanonicalexamples}. However, representing contextual meaning as additive combinations of fixed sense vectors may limit expressivity when sense interactions are non-linear.
\paragraph{Semantic Concept Integration.}
While Backpack Language Models decompose lexical inputs, \citet{tack2025llmpretrainingcontinuousconcepts} introduce CoCoMix to enforce decomposability at the level of higher-level semantic concepts. Instead of operating solely over discrete token representations, CoCoMix integrates SAE into pretraining, training the model to predict continuous concept representations alongside next-token probabilities. These predicted concept vectors are interleaved with hidden states, encouraging explicit reasoning over disentangled semantic features during generation. By treating interpretable concepts as structured components of the forward pass, CoCoMix enables targeted control over generation while preserving output coherence, at the cost of introducing additional training structure and reliance on the quality of concept representations.

\subsection{Explicit Modularization} \label{subsec:moe}

\begin{figure*}
    \centering
    \includegraphics[width=0.98\linewidth]{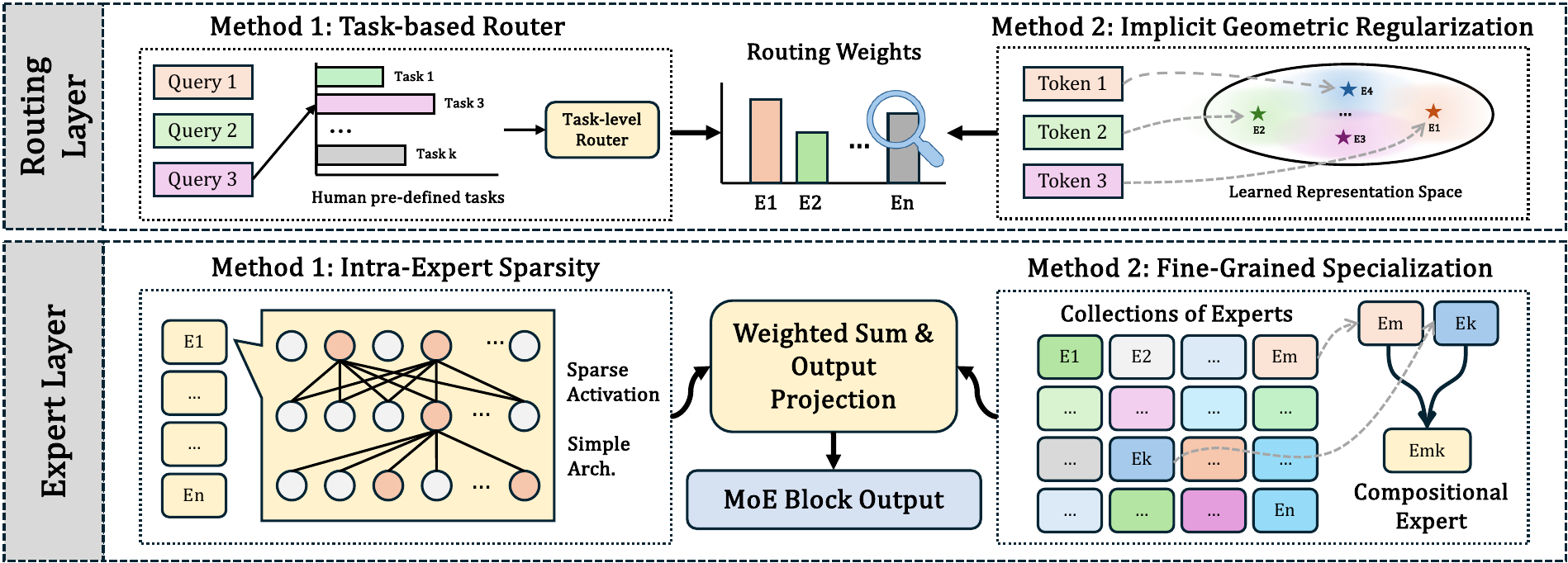}
    \caption{\textbf{Architectural strategies for intrinsically interpretable MoEs.} We distinguish between methods enforcing intra-expert sparsity, fine-grained decomposition, and semantically aligned routing.}
    \label{fig:moe}
\end{figure*}

In practice, explicit modularization is most often realized through MoE architectures. While standard MoE models are primarily designed for scalability, their expert representations and routing mechanisms are typically optimized for load balancing rather than semantic transparency \citep{DBLP:journals/jmlr/FedusZS22}. Recent work revisits MoE design with interpretability as a central goal. We organize these methods into three architectural strategies, illustrated in \Cref{fig:moe}, which we discuss in turn below.

\paragraph{Enforcing Intra-Expert Sparsity and Simplicity.}
A direct approach is to constrain the experts themselves. One strategy replaces smooth activations (e.g., GeLU) with hard thresholds like ReLU. For instance, MoE-X \citep{DBLP:conf/icml/YangVKWDB025} uses this to enforce sparsity on hidden states, helping to disentangle features. A parallel strategy simplifies the expert architecture. Methods like MoV and MoLORA \citep{DBLP:conf/iclr/ZadouriUAELH24} replace full MLPs with lightweight vectors or low-rank adapters. While primarily efficient, these linear or low-rank experts are also much easier to analyze than deep, non-linear MLPs. However, despite simplifying expert internals, these models still rely on routing and sparse expert selection, and therefore remain sensitive to load imbalance during training.

\paragraph{Architecting for Fine-Grained Decomposition.}
Another strategy seeks monosemanticity by scaling the number of experts to match the number of features. Building on tensor decomposition methods like MPO-MoE \citep{gao-etal-2022-parameter}, architectures such as MONET \citep{DBLP:conf/iclr/ParkJKK25} and MxD \citep{oldfield2025interpretabilitysacrificefaithfuldense} use product key composition and flexible tensor factorization. These techniques construct hundreds of thousands of fine-grained sublayers from a compact parameter set, effectively treating the MoE layer as a sparse dictionary of specialized linear transformations. Despite their improved granularity, the expansion of the expert space amplifies routing sensitivity, making training vulnerable to routing imbalance and expert under-utilization.

\paragraph{Designing Semantically Aligned Routing Policies.}
While early efforts simply mapped experts to languages \citep{zhao2024sparse}, recent work distinguishes between explicit structural alignment and implicit geometric regularization. In the explicit paradigm, models like Task-Based MoE \citep{pham2023taskbasedmoemultitaskmultilingual} and THOR-MoE \citep{liang2025thormoehierarchicaltaskguidedcontextresponsive} directly integrate task context into the router, whereas Apollo-MoE \citep{DBLP:conf/iclr/ZhengWL0ZW25} organizes experts by linguistic families. In the implicit paradigm, researchers enforce constraints on the routing space itself: RoMA \citep{li2025routingmanifoldalignmentimproves} aligns routing manifolds with task embeddings, and USMoE \citep{do2025unifiedsparsemixtureexperts} reframes selection as a linear programming problem. Supporting these directions, recent analyses confirm that routing decisions follow distinct layer-wise patterns and can be predictably steered to alter model behavior \citep{DBLP:journals/corr/abs-2510-04694,DBLP:conf/iclr/ZhengWL0ZW25}.
%\textcolor{red}{[add one or two lines of limitation if needed]}

\subsection{Latent Sparsity Induction}
\label{subsec:latent sparsity induction}

Unlike explicit modular architectures such as Mixture-of-Experts, latent sparsity induction aims to encourage modular and interpretable structure to \emph{emerge} within otherwise standard Transformer architectures. Rather than prescribing a fixed decomposition, these methods introduce inductive biases that promote selective activation and reduce superposition, allowing the model to organize its computation into sparse, task-specific subcircuits.

\paragraph{Enforcing Weight Sparsity.} %\textcolor{red}{[check, I merged enforcing sparsity and bridging sparse and dense models]}
A primary approach to latent sparsity induction enforces sparsity directly at the level of model parameters. The underlying hypothesis is that polysemanticity arises from dense connectivity, where individual neurons participate in many unrelated computations. To address this, recent work trains Transformer models with strong sparsity constraints
% , such as $L_0$ regularization 
\citep{gao2025weightsparsetransformersinterpretablecircuits}, imposing sparsity throughout optimization rather than post-hoc pruning, forcing the model to allocate its limited connections more selectively and reducing feature superposition \citep{elhage2022toymodelssuperposition}.
A direct consequence of sparse training is the emergence of compact and interpretable computational circuits.

While weight-sparse models offer strong interpretability benefits, they are often inefficient on current hardware. \citet{gao2025weightsparsetransformersinterpretablecircuits} train sparse and dense models jointly, coupled through linear mappings between their representations, making interpretable features discovered in the sparse model usable to explain or annotate the latent space of the dense model.
% , combining the efficiency of dense architectures with the structural clarity induced by sparsity.
However, \citet{gao2025weightsparsetransformersinterpretablecircuits} note that enforcing weight sparsity introduces a capability-interpretability trade-off and remains difficult to scale.

\paragraph{Conditional Activation via Gated Architectures.}
Latent sparsity can also be induced at the level of activations rather than weights. Gated architectures, such as GLU and SwiGLU \citep{DBLP:conf/icml/DauphinFAG17,shazeer2020gluvariantsimprovetransformer}, introduce conditional computation within Transformer feed-forward layers. Unlike standard pointwise activations (e.g., ReLU \citep{DBLP:journals/jmlr/GlorotBB11} or GeLU \citep{hendrycks2023gaussianerrorlinearunits}), GLUs compute an element-wise product between a value projection and a learned gate:
\[
\text{GLU}(x) = (xW) \odot \sigma(xV)
\]
Here, the gating term selectively suppresses or amplifies features on a per-input basis, effectively routing information through different subspaces. While GLUs do not enforce strict sparsity, their conditional structure encourages selective pathway activation, reducing entanglement and promoting emergent modularity. This form of activation-level sparsity complements weight-sparse approaches by enabling input-dependent specialization without explicitly defined modules.

%% file: future_direction.tex
\section{Open Challenges and Future Directions}

Despite recent progress, intrinsic interpretability for LLMs remains an open and rapidly evolving research area. We highlight several key challenges and promising directions for future work.

\paragraph{Defining and Evaluating Intrinsic Interpretability.}
A central challenge is the lack of rigorous and widely accepted definitions and evaluation metrics for intrinsic interpretability \cite{doshivelez2017rigorousscienceinterpretablemachine,DBLP:journals/cacm/Lipton18}. Although intrinsic approaches aim to align model structure with explanation, it remains unclear how to quantitatively assess the quality, completeness, or usability of such explanations. Existing evaluations rely primarily on proxy measures such as sparsity, modularity, or disentanglement, which do not reliably reflect human interpretability or task relevance. In particular, structural properties like sparsity do not guarantee semantic clarity, as features may remain polysemantic or lack stable human interpretable meaning \citep{elhage2022toymodelssuperposition}. Moreover, without principled verification, intrinsically generated explanations may appear plausible while failing to faithfully represent the model’s true reasoning \citep{DBLP:conf/nips/TurpinMPB23,singh2024rethinkinginterpretabilityeralarge}. Developing evaluation frameworks that balance faithfulness, human comprehensibility, and downstream utility therefore remains an important open problem.

\paragraph{Balancing Interpretability and Expressivity.}
Although recent work suggests that interpretability and performance need not be mutually exclusive, intrinsic constraints may still limit model expressivity or generalization in practice \citep{gao2025weightsparsetransformersinterpretablecircuits, DBLP:conf/iclr/SunOUW25}. Understanding when and how architectural biases such as modularity, sparsity, or concept alignment improve or hinder learning remains an open question. Future research should aim to characterize these trade-offs more precisely, identifying regimes in which intrinsic interpretability enhances robustness and generalization rather than restricting model capacity.

\paragraph{Scalability to Large-Scale Language Models.}
Most intrinsically interpretable architectures have so far been evaluated only at small or moderate scales \cite{DBLP:conf/iclr/ParkJKK25,DBLP:conf/icml/TamkinTG24,tack2025llmpretrainingcontinuousconcepts}. 
Extending these designs to large language models with billions of parameters introduces additional challenges, including increased routing complexity, memory overhead, and optimization instability. Demonstrating that intrinsic interpretability can be preserved at scale therefore remains a critical step toward practical deployment.
% without relying on post-hoc approximations.

\paragraph{Training Efficiency and Optimization Stability.}
Intrinsic interpretability often introduces additional constraints or architectural components, such as sparse activations, modular routing, or complex functional parameterizations, which can complicate optimization and increase training cost \citep{gao2025weightsparsetransformersinterpretablecircuits,jin2025megascalemoelargescalecommunicationefficienttraining,DBLP:conf/iclr/LiuWVRHS0T25}. Improving the efficiency and stability of training intrinsically interpretable models is therefore an important direction for future work. This includes developing better optimization strategies, regularization schemes, and hardware-aware implementations that make intrinsic designs competitive with standard dense architectures.

\paragraph{Complementarity with Post-hoc Analysis.}
Although intrinsic and post hoc interpretability are often framed as distinct paradigms, they need not be mutually exclusive. Post-hoc tools can act as diagnostic instruments for validating and stress testing intrinsically interpretable models, while intrinsic structural constraints can in turn improve the faithfulness of post-hoc analyses. Moreover, insights derived from post-hoc methods can inform intrinsic design, for example by guiding concept discovery, feature selection, or module construction \cite{tack2025llmpretrainingcontinuousconcepts}. Developing principled frameworks that integrate intrinsic architectures with post-hoc analysis therefore represents a promising direction for building more transparent and reliable LLMs.

\section{Conclusion}

In this paper, we present a comprehensive survey of intrinsic interpretability for large language models. We first clarify the key distinctions between intrinsic interpretability and post-hoc explanation methods, highlighting their conceptual differences and respective strengths. We then categorize and analyze existing approaches through the lens of five core design principles: functional transparency, concept alignment, explicit modularization, latent sparsity induction, and representational decomposability. In addition, we synthesize recent advances across model architectures and training strategies, and identify five key challenges and future directions for intrinsically interpretable model design. Our goal is to provide a structured and accessible resource for researchers interested in building transparent, interpretable-by-design LLMs. To aid in this effort, we create a continually updated paper list in a GitHub repository as follows: \href{https://github.com/PKU-PILLAR-Group/Survey-Intrinsic-Interpretability-of-LLMs}{\faGithub \ Survey-Intrinsic-Interpretability-of-LLMs}.

\section*{Limitations}

While this survey aims to provide a comprehensive overview of intrinsic interpretability for large language models, several limitations regarding its scope should be noted. The field is evolving rapidly, and despite efforts to incorporate work up to late 2025, new architectures and training techniques continue to emerge. Accordingly, this survey reflects a snapshot of the current literature and may not cover unpublished or concurrent preprints. To balance breadth and clarity, we emphasize unifying principles rather than exhaustive technical detail for individual methods.

\section*{Acknowledgement}

This work was supported in part by the Beijing Major Science and Technology Project under Contract No. Z251100008125054. This work was supported by the Beijing Academy of Artificial Intelligence (BAAI).